\pdfoutput=1

\documentclass[11pt]{article}

\usepackage{acl2021}

\usepackage{times}
\usepackage{latexsym}
\usepackage[T1]{fontenc}
\usepackage[utf8]{inputenc}
\usepackage{microtype}

\usepackage{graphicx}
\usepackage{algorithm}
\usepackage{algpseudocode}
\usepackage{multirow}
\usepackage{booktabs}
\usepackage{colortbl}
\usepackage{xcolor}
\usepackage{amsmath}
\usepackage{mathtools}
\usepackage{subfigure}
\usepackage{adjustbox}
\usepackage{amssymb}
\usepackage{tikz}
\usepackage{amsmath}
\usepackage{url}

\usepackage{microtype}

\title{
    Challenges in Generalization in Open Domain Question Answering
}

\author{
    Linqing Liu$^{\dagger{}}$ \hspace{0.15cm}
    Patrick Lewis$^{\ddagger{}\dagger{}}$ \hspace{0.15cm} 
    Sebastian Riedel$^{\ddagger{}\dagger{}}$ \hspace{0.15cm}
    Pontus Stenetorp$^{\dagger{}}$ \hspace{0.15cm} \\
    $^{\dagger{}}$University College London \hspace{0.3cm} $^{\ddagger{}}$Facebook AI Research\\
    \texttt{\{linqing.liu,patrick.lewis,s.riedel,p.stenetorp\}@cs.ucl.ac.uk} \\
    \\
}

\date{}

\newcommand{\zeroshot}{novel-entity}
\newcommand{\capzeroshot}{Novel-entity}

\begin{document}
\maketitle
\begin{abstract}
Recent work on Open Domain Question Answering has shown that there is a large discrepancy in model performance between \textit{novel} test questions and those that largely overlap with training questions.
However, it is unclear which aspects of novel questions make them challenging. 
Drawing upon studies on systematic generalization, we
introduce and annotate questions according to three categories that measure different levels and kinds of generalization:
\textit{training set overlap}, \textit{compositional generalization~(comp-gen)}, and \textit{novel-entity generalization~(novel-entity)}.
When evaluating six popular parametric and non-parametric models, we find that for the established Natural Questions and TriviaQA datasets, even the strongest model performance for comp-gen/\zeroshot{} is 13.1/5.4\% and 9.6/1.5\% lower compared to that for the full test set -- indicating the challenge posed by these types of questions.
Furthermore, we show that whilst non-parametric models can handle questions containing novel entities relatively well, they struggle with those requiring compositional generalization.
Lastly, we find that key question difficulty factors are: cascading errors from the retrieval component, frequency of question pattern, and frequency of the entity.
\end{abstract}

\section{Introduction}
Over the last few years we have seen model innovations improving on standard natural language processing~(NLP) benchmarks across the board~\cite{devlin2019bert, raffel2020exploring, lewis2020bart}.
However, it is still clear that we are yet to obtain generalizable language understanding, as recent work has found that adversarial \cite{jia2017adversarial, mudrakarta2018did, belinkov2018synthetic} and out-of-distribution samples \cite{talmor2019multiqa, elsahar2019annotate, mccoy-etal-2020-berts} remain challenging for existing models across numerous tasks.
\begin{figure}
\centering
\includegraphics[width=\linewidth]{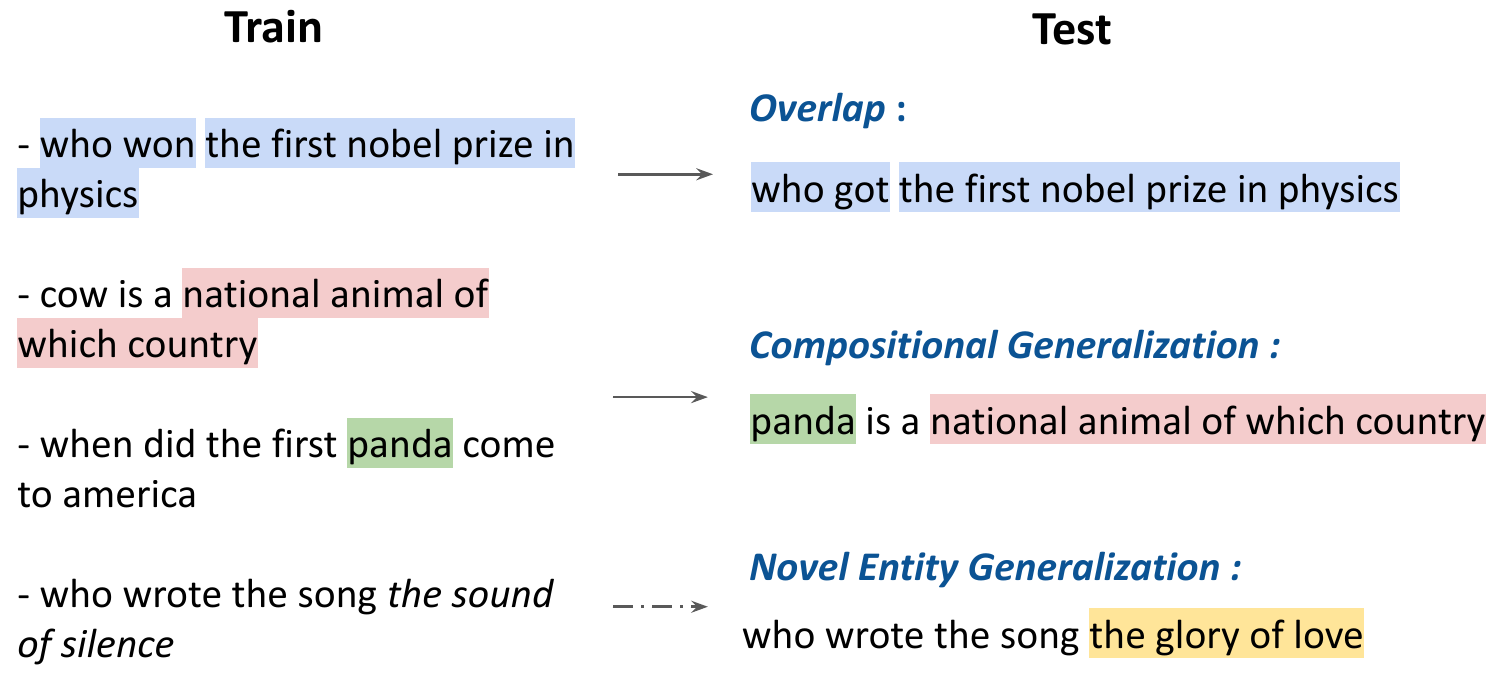}
\caption{
    Questions categorized according to their relation to the training set:
    1) \textit{Overlap}: there exists a paraphrase of the question in the training set.
    2) \textit{Compositional}: all individual facts and the structure of the question has been observed across several questions in the training set -- but not the given composition.
    3) \textit{\capzeroshot{}}: the question contains at least one entity (marked here with yellow) not present in the training set.
}
\label{fig:def_gen}
\end{figure}

Open-domain question answering~(ODQA), which aims to answer factoid questions without any given context, is a task that has been receiving increasing attention in the community~\cite{chen2017reading, lee2019latent, karpukhin2020dense, izacard2021leveraging, min2021neurips}.
However, recent work has shown that there is a large discrepancy in model performance between questions and answers observed at train time and \emph{novel} questions and answers -- even if they are derived from the same distribution~\cite{lewis2021question}.
This raises the question:
``What are the aspects of these novel questions that make generalization challenging?'' which we seek to explore in this paper.
%
%

In work on systematic generalization~\cite{bahdanau2018systematic, lake2018generalization, ruis2020benchmark}, it is argued that even though a model has only observed a very small subset of all possible combinations of facts during training time, a good model should be able to generalize to all possible combinations of facts at test time.
We draw upon these ideas to study generalization for ODQA and define the following three categories to support our investigation: \textit{training set overlap}, \textit{compositional generalization}, and \textit{\zeroshot{} generalization}.
See Figure~\ref{fig:def_gen} for definitions and examples.
Our categorization breakdown is motivated by how they capture different levels of generalization: \emph{overlap} requiring no generalization beyond recognizing paraphrases, \emph{comp-gen} requiring generalization to novel compositions of previously observed entities and structures, and \emph{\zeroshot{}} requiring generalization to entities not present in the training set.
It is worth noting that we explicitly study in-distribution generalization rather than out-of-distribution generalization~(such as cross-domain generalization~\citep{fisch2019mrqa}), as we will later demonstrate that even in-distribution generalization poses a major challenge for existing approaches.

We decompose and manually annotate three previously introduced ODQA datasets~(Natural Questions~\citep{lee2019latent}, TriviaQA~\citep{joshi2017triviaqa}, and WebQuestions~\citep{berant2013semantic}).
Following this, we evaluate six recently proposed non-parametric and parametric ODQA models and analyze their performance, using both aggregate metrics and a breakdown according to our proposed categories.
Non-parametric and parametric models differ in their access to information: the former has no access to any external context or knowledge, whereas the latter is provided relevant information alongside the question~\cite{roberts2020much}.

One potential source of difficulty could be the question structure itself and as a byproduct of our decomposition approach we are able to derive a high-level \emph{question pattern} for each question. We find a strong positive correlation between the pattern frequency in the training set and test accuracy. 
We then study how non-parametric models handle the comp-gen and \zeroshot{} subsets respectively, since the performance on them is significantly worse than on the overlap subset. 
For \emph{comp-gen} questions, perhaps surprisingly, we find that the frequency of entities mentioned in a question is strongly \emph{negatively} correlated with test accuracy. 
For \emph{\zeroshot{}} questions, when we replace novel entities in the question and its support passages with entities seen in the training set the performance remains largely unchanged;
we thus hypothesize that specific unseen entities are not the main bottleneck for model performance but rather a failure of the model to generalise compositionally.
Aside from questions, we further analyze the retrieved passages and find the retrieval accuracy is equally lacking for the \emph{comp-gen} and \emph{\zeroshot{}} subsets, at $\sim75\%$ for top-20 accuracy.
We also observe that many of the passages that do contain the correct answer lack sufficiently informative contexts for the question anchor words for the reader model to be able to locate it, indicating a need to either improve the reader models ability reason over multiple passages or the retriever model to provide passages with richer contexts.

To conclude, our key contributions are as follows: 
1) We provide the first detailed study on generalization for ODQA, based on categories that measure different levels and kinds of generalization, that we use to annotate three previously proposed ODQA datasets
\footnote{Our data and code are available at \url{https://github.com/likicode/QA-generalize}}.
2) We show that for novel questions, non-parametric models handle novel question entities comparatively well, while they struggle to perform compositional generalization. 
3) We demonstrate and quantify key factors that impact model generalization performance, which we believe will show the direction for future research towards more robust and generalizable ODQA models.

\newcommand{\annotate}[1]{$\text{\underline{\textit{#1}}}$}
\section{Dataset Construction}

In this section, we describe how we process and annotate ODQA datasets to enable us to investigate generalization.

\subsection{Question Decompostition}
\label{sec:question_decomp}
To study the compositional and \zeroshot{} generalization of questions, we follow \citet{keysers2019measuring} and propose to view each question as being composed of primitive elements~(atoms).
Consider the question \textit{``Who got the first Nobel Prize in Physics?''}.
The atoms intuitively correspond to the modifier or adjunct of the predicate ``who'', predicate ``got'' and the entity ``first nobel prize in physics''.
The combination of these atoms cover the main semantics of the question.

The way we measure generalization necessarily depends on how we break down the questions into atoms. 
Following manual analysis of questions from three popular ODQA datasets, we developed the following decomposition strategy to obtain atoms which cover all the desired question semantics. These are:
question words, verbs, Wikipedia named entities (\textit{wiki\_entities}), and finally, other arguments (\textit{other\_args}) which correspond to other relevant aspects of the question.
We explicitly extract wiki\_entities since they leverage crucial semantics in factoid questions and other\_args define essential details surrounding wiki\_entities. 

In order to automatically decompose questions, we first use an off-the-shelf semantic role labeling (SRL) model \cite{shi2019simple} to produce predicate-argument structures for each question. This provides us with the verb (i.e. the predicate), and semantic arguments. The question word is trivially obtained by identifying WH-words. 
We apply an off-the-shelf entity linking model~\cite{li2020efficient} to obtain the wiki\_entities in the question.
Finally, other\_args are the SRL arguments which remain after we filter out arguments corresponding to wiki\_entities.
An example question decomposition is illustrated in Figure~\ref{fig:decomp_question}. More details are included in  Appendix~\ref{appendix:question_decomp}.

\begin{figure}
\centering
\includegraphics[width=\linewidth]{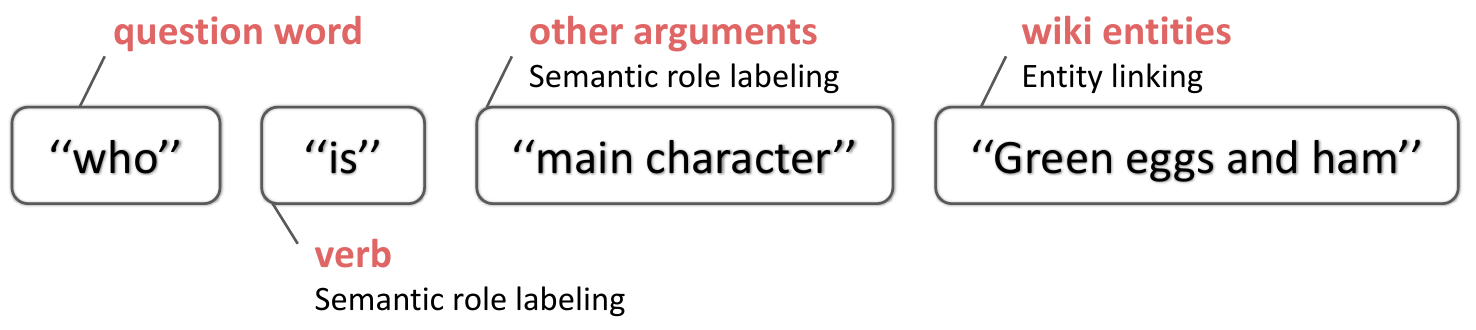}
\caption{
    Example decomposition for the question \textit{``Who is the main character in Green eggs and ham?''}
}
\label{fig:decomp_question}
\end{figure}

\begin{table}
\centering
\resizebox{1.0\columnwidth}{!}{
\begin{tabular}{lrrr}
\toprule
Group     & Natural Questions & WebQ & TriviaQA \\
\midrule
Overlap   &  837  & 501 &   458       \\
Comp-gen  &  1,105  & 512  &  475      \\
\capzeroshot{} &  597  & 640 & 456 \\
\bottomrule
\end{tabular}}
\caption{
    Number of questions for each generalization subset for the three datasets' test sets
}
\label{tab:statistics}
\end{table}

\subsection{Generalization Category Definitions}
\label{sec:gen_define}
Based on the question decomposition, we define three generalization categories for ODQA datasets.
We denote $S_{q}$ as the set of the decomposed atoms of question $q$ and $C_{Q}$ as the complete set of decomposed atoms for all the questions in dataset $Q$.
Our category subsets are then defined as:

\begin{itemize}
    \item $Q_{\text{overlap}} \triangleq \{ q \in Q_{\text{test}} \mid \exists \, q' \in Q_{\text{train}}, S_{q} \subseteq S_{q'} \}$
    
    \item $Q_{\text{comp\_gen}} \triangleq \{ q \in Q_{\text{test}} \mid
    \exists \, q'_{1}, q'_{2}, ..., q'_{k} \in Q_{\text{train}}, \, S_{q} \subseteq \bigcup_{i=1}^{k} S_{q'_{i}}, \, S_{q} \not\subseteq S_{q'_{i}}
    \}$
    
    \item $Q_{\text{novel\_entity}} \triangleq \{ q \in Q_{\text{test}} \mid 
    \exists \, s \in S_{q}, s \notin C_{\text{train}}
    \}$
\end{itemize}

For overlap test question, there exists a training question where they have the same decomposed atoms or are subset of them; for comp\_gen test question, its decomposed atoms are fully covered by the training set~(a subset of the union of multiple training questions atoms), but not in one particular training question; and for \zeroshot{} test question, there exist wiki\_entities not present in the training set.

\begin{table*}
\resizebox{\textwidth}{!}{
\begin{tabular}{lllc}
\toprule
\multicolumn{1}{l}{Group}  & Test question         & \begin{tabular}[c]{@{}l@{}}Paired training question for annotator\end{tabular} & \multicolumn{1}{l}{Label} \\ \hline
\multirow{2}{*}{Overlap}   & who got the first nobel prize in physics           & who won the first nobel prize in physics                        & T \\
& whens the last time the patriots played the eagles & when did the philadelphia eagles last win the super bowl & F  \\ \hline
\multirow{2}{*}{Comp-gen}  & when is the next scandal episode coming out        & when is next fairy tail episode coming out & T\\
& what is the corporate tax rate in great britain    & what is the rate of corporation tax in uk      & F  \\ \hline
\multirow{2}{*}{\capzeroshot{}} & who wrote the song the \textit{glory of love}   & who sang \textit{guilty of love} in the first degree & T \\
& who sings \textit{too much time on my hands} lyrics & who sings \textit{i’ve got too much time on my hands}   & F  \\
\bottomrule                      
\end{tabular}}
\caption{
    Example of questions from Natural Questions~(see Appendix~\ref{appendix:question_examples} for examples from the other two datasets) for human verification and their respective annotated labels (T for True and F for False).
    }
\label{tab:annot}
\end{table*}

\subsection{Question Categorization and Human Verification}
\label{sec:manual_verify}
With the decomposed atoms for all questions, we first categorize the test questions into overlap, comp-gen, and \zeroshot{} categories based on the definitions of each generalization category.
We optimize the selection criteria to cover as many eligible candidates for each category as possible.
Further details can be found in Appendix~\ref{appendix:question_collection}.

As our test set subsets are obtained automatically, we need to perform manual human verification to ensure that they are of high enough quality to draw empirical conclusions.
To do this, we employ four expert annotators and use the following annotation process for each of the respective categories.
\textit{Overlap:} Annotators are shown $q_{\text{test}}$ and the training questions with the highest degree of character-level overlap.
If any of these questions are a paraphrase of $q$, the annotator will mark $q_{\text{test}}$ as an \emph{overlap} question.
\textit{Comp-gen: } $q_{\text{test}}$ is presented to the annotators along with the training questions with the highest degree of word overlap.
Annotators then verify that the test question is truly a compositional generalization and not a paraphrase of any of the given training questions.
\textit{\capzeroshot{}: } 
Annotators need to: 1) Verify 
that the wiki\_entities identified by the entity-linking model are indeed wiki entities.
2) Verify that the entities in $q_{\text{test}}$ are not present among a set of questions from the training set whose entities have a high degree of character-level overlap with the entities in $q_{\text{test}}$.
Statistics for the annotated category subsets are summarized in Table~\ref{tab:statistics}, examples are shown in Table~\ref{tab:annot}, and additional details are covered in Appendix~\ref{appendix:subset_detail}.

\section{Experiment}
\subsection{Datasets}

We analyse three widely used ODQA datasets, each one is briefly introduced as follows:

\paragraph{Open Natural Questions (NQ)} is an open-domain variant of Natural Questions~\cite{kwiatkowski-etal-2019-natural} introduced by \citet{lee2019latent}. This dataset consists of questions mined from Google search logs, with answers annotated as short spans of text in Wikipedia articles by crowd-workers. The NQ questions are generally simple, short, and \emph{information-seeking}, as the questioner is unlikely to have known the question's answer when they formulated it. 
It consists of 79,168 train, 8,757 dev, and 3,610 test question answer pairs.

\paragraph{TriviaQA} \cite{joshi2017triviaqa} consists of questions and answers which were obtained by scraping trivia websites. 
TriviaQA questions are generally less information-seeking than those in NQ, and exhibit substantial syntactic and lexical variability. 
We use the open domain splits which contains 78,785 train, 8,837 dev, and 11,313 test question answer pairs \cite{lee2019latent}.
Answers in TriviaQA are Wikipedia entities, and any alias of the answer entity is considered a correct answer. We randomly sampled and annotated 2,000 questions from the test set for our analyses.
%
%

\paragraph{WebQuestions} \cite{berant2013semantic} consists of questions that were collected by performing a breadth-first search using the Google Suggest API.
The questions in WebQuestions resemble those in NQ, but are generally shorter and simpler and demonstrate less variability. WebQuestions' answers are Freebase~\cite{bollacker2008freebase} entities, annotated by crowdworkers.
It contains 3,778 train and 2,032 test questions.

\begin{table*}
\centering
\begin{adjustbox}{width=1\textwidth}
\begin{tabular}{l|l|rrrr|rrrr|rrrr} 
\toprule
\multicolumn{2}{c|}{\multirow{2}{*}{Model}}      & \multicolumn{4}{c|}{Natural Questions~}  
& \multicolumn{4}{c|}{TriviaQA}              & \multicolumn{4}{c}{WebQuestions}                   \\ 
\multicolumn{2}{l|}{} & Total & Overlap & \begin{tabular}[c]{@{}l@{}}Comp\\-gen\end{tabular} & \begin{tabular}[c]{@{}l@{}}Novel\\-entity\end{tabular}
& Total & Overlap & \begin{tabular}[c]{@{}l@{}}Comp\\-gen~\end{tabular} & \begin{tabular}[c]{@{}l@{}}Novel\\-entity\end{tabular} 
& Total & Overlap & \begin{tabular}[c]{@{}l@{}}Comp\\-gen\end{tabular} & \begin{tabular}[c]{@{}l@{}}Novel\\-entity\end{tabular}  \\ 
\midrule
\multicolumn{1}{l}{\multirow{4}{*}{Non-parametric}} & RAG & 44.49 & 75.75 & 30.41 & 37.69 &
56.83 & 87.12 & 47.58 & 47.81 &
45.52 & 80.64 & 33.40 & 31.88 \\
\multicolumn{1}{l}{} & FiD & \textbf{53.13} & 78.85   & \textbf{40.00}  & \textbf{47.74} &
\textbf{67.69} &\textbf{90.39} & \textbf{58.10} & \textbf{66.23} &
- & - & - & - \\ 
\multicolumn{1}{l}{} & DPR & 41.27 & 71.33   & 25.88  & 33.84 &
57.91 & 82.31 & 46.11 & 58.99 &
42.42 &  73.45 & 31.05 & 31.25 \\
\multicolumn{1}{l}{}  & RePAQ        & 47.26 & 78.61 & 34.21 & 36.85&
52.06 & 89.08 & 42.95 & 38.38 & 
-& - & -& -\\
\midrule
\multicolumn{1}{l}{\multirow{2}{*}{Parametric}} & T5-11B+SSM & 36.59 & \textbf{81.48}   & 17.47 & 12.56 & - & - & - & - & 44.69 & 81.24 & 35.35  & 25.78\\ 
\multicolumn{1}{l}{}                    & BART & 26.54 & 76.34 & 5.88 & 3.35 &
26.78 & 78.38 & 11.37 & 10.09 & 
27.41  & 70.46 & 13.28& 8.75 \\
\bottomrule
\end{tabular}
\end{adjustbox}
\caption{
    Exact Match scores for each model. ``Total'' refers to the overall performance on the full test set. ``Overlap'', ``Comp-gen'', and ``\capzeroshot{}'' refers to the model performance on the respective subset.
}
\label{tab:main_result}
\end{table*}

\subsection{Baseline Models}
\paragraph{Non-parametric models} mostly adopt a retrieve-and-read framework, retrieving relevant Wikipedia documents for the given question, and then produce the final answer conditioned on these documents.
%
We consider two generative reader models: Retrieval-Augmented Generation~\cite[RAG,][]{lewis2020retrieval}, and  Fusion-In-Decoder~\cite[FiD,][]{izacard2021leveraging}. RAG combines a DPR~\cite{karpukhin2020dense} dense retriever with a BART~\cite{lewis2020bart} generator, which are jointly fine-tuned end to end. 
%
FiD is a pipeline approach which uses DPR to retrieve a set of documents, and the decoder attends over all encoded document representations to generate the final answer.
As an extractive reader model we use the reader component from DPR \cite{karpukhin2020dense}. 
%
It extracts answer span from the highest-scoring document ranked from a passage selection model.
We also include RePAQ \cite{lewis2021paq}, a QA-pair retriever which does \textit{not} follow the retrieve-and-read paradigm. 
It retrieves QA-pairs from PAQ, a large resource of 65M automatically-generated QA-pairs, returning the answer of the most relevant QA-pair.

\paragraph{Parametric models} are directly trained with QA pairs without access to an external corpus and thus store the required knowledge in its entirety in the model parameters.
For our analyses, we include a BART-large model \cite{lewis2020bart} and a more powerful T5-11B model \cite{roberts2020much}.
They are both trained with questions as input and output question-answer pairs. 

\subsection{Model Category Analysis}

Table~\ref{tab:main_result} shows the Exact Match scores for models on our test set splits. 

\paragraph{Non-parametric models on \zeroshot{} questions} 
For the non-parametric models, EM scores on \emph{\zeroshot{}} questions are relatively close to their overall total scores, with an average drop by 6.5\% and 3.1\% on NQ and TriviaQA respectively, with the exception of WebQuestions.
The questions in WebQuestions only contain a single entity, which also tend to be high frequency entities.
However, due to the very small size of the WebQuestions training set, many of these questions are considered to be in the \zeroshot{} subset, despite containing relatively frequent entities, which, with a larger training set, would likely be classified as comp-gen questions, 
querying various relations regarding known entities.

\paragraph{Non-parametric models on comp-gen questions}
Surprisingly, the performance of all non-parametric models degrades significantly on the \emph{comp-gen subset} (drop by 
14.2\% on NQ, 10.2\% on TriviaQA and 11.7\% on WebQuestions). 
This finding suggests that non-parametric models struggle to perform compositional generalization, whereas they handle novel question entities comparatively well. We investigate this finding in greater detail in Section~\ref{sec:discuss}.

\paragraph{Parametric models on \zeroshot{} and comp-gen questions}
parametric model performance drops significantly on both comp-gen and \zeroshot{} subsets, but they achieve relatively higher EM scores on comp-gen questions.
This indicates that \zeroshot{} questions are more challenging for parametric models.
This makes intuitive sense, since, for entities not seen during training, parametric models will struggle to ``know" enough about the entity to generate a correct answer.
In such cases, we find evidence that parametric models often resort to generating answers from superficially similar training questions, with 63.2\% and 53.3\% of answer predictions also occurring in the training data for $\text{T5-11B+SSM}$ on NQ for comp-gen and \zeroshot{} questions respectively.

\paragraph{Implications for modeling}
\label{para:answer_freq}
Among the non-parametric models, FiD achieves the highest EM scores for both comp-gen and \zeroshot{} questions. FiD aggregates multiple passages together when generating  answers.
In contrast, the extractive DPR reader only uses the highest-scoring passage to extract the final answer. 
Based on observations from the experiment in Appendix~\ref{appendix:implication_for_modeling},
we hypothesize that the NQ FiD model adopts a strategy similar to a reranker, and extracts an answer from the highest latently-relevant document.

Although without access to external knowledge but only automatically-generated QA-pairs in advance when answering questions, RePAQ still achieves higher or comparable performance as retrieve-and-read model RAG and DPR.
It indicates that generating, storing and retrieving questions is a valid path in terms of model generalization.

Parametric models perform significantly worse compared to non-parametric models. 
BART struggles to answer any novel questions correctly, while $\text{T5-11B+SSM}$ performs better due to much larger capacity. 
\citet{petroni2019language} demonstrate that language models are able to recall factual knowledge without any fine-tuning and can somewhat function as an unsupervised ODQA system. 
However, our experiments suggest that, large-scale language models (when fine-tuned to directly answer questions using a set of training QA pairs) struggle to answer questions about low frequency entities and relations, similar to the findings of \citet{kassner2020pretrained} and \citet{dufter2021static}.

\paragraph{Additional observations} 
All models perform significantly higher on overlap questions, consistent with the findings of \citet{lewis2021question}. Parametric models with more parameters are the most effective at rote-memorizing training questions, and $\text{T5-11B+SSM}$ even outperforms the non-parametric models on NQ and WebQuestions.

\section{How Do Non-parametric Models Generalize?}
\label{sec:discuss}

\begin{table}
\centering
\resizebox{\columnwidth}{!}{
\begin{tabular}{lcccc}
\toprule
NQ      & Total & Overlap & Comp-gen & \capzeroshot{}  \\
\midrule
Top-20  & 80.1  & 89.5    & 74.7     & 75.4       \\
Top-100 & 86.1  & 92.0    & 82.4     & 83.1       \\
\bottomrule
\end{tabular}}
\caption{Top 20 and Top 100 retrieval accuracy on NQ test set for the DPR retriever.}
\label{tab:retriever_result}
\end{table}

Experimental results show that the performance of non-parametric models degrades significantly on the comp-gen subsets across all datasets. In this section, we would like to examine what the underlying challenge is for these questions.
We focus on the NQ dataset as it has the largest annotated test set among three datasets.

Table~\ref{tab:retriever_result} shows the
top-$k$ retrieval accuracy -- which is the number of questions for which at least one passage of the top-$k$ retrieved passages contains the gold answer.
The difference in retrieval accuracy between comp-gen and \zeroshot{} splits is relatively small~($< 1\%$), but is significantly lower than the overlap subset results. 
This indicates that the retriever performance is a confounding factor for the overall performance of comp-gen and \zeroshot{} questions.
Solely improving the retriever would benefit the model greatly for the subsets requiring generalization.
Allowing us to study the reader model in isolation, for the remainder of our analysis we will only use the subset of questions for which there is at least one support passage that contains the gold answer.

\subsection{Effects of Question Pattern Frequency}
\label{sec:question_pattern}

\begin{figure}
\centering
\subfigure{
\includegraphics[width=.78\columnwidth]{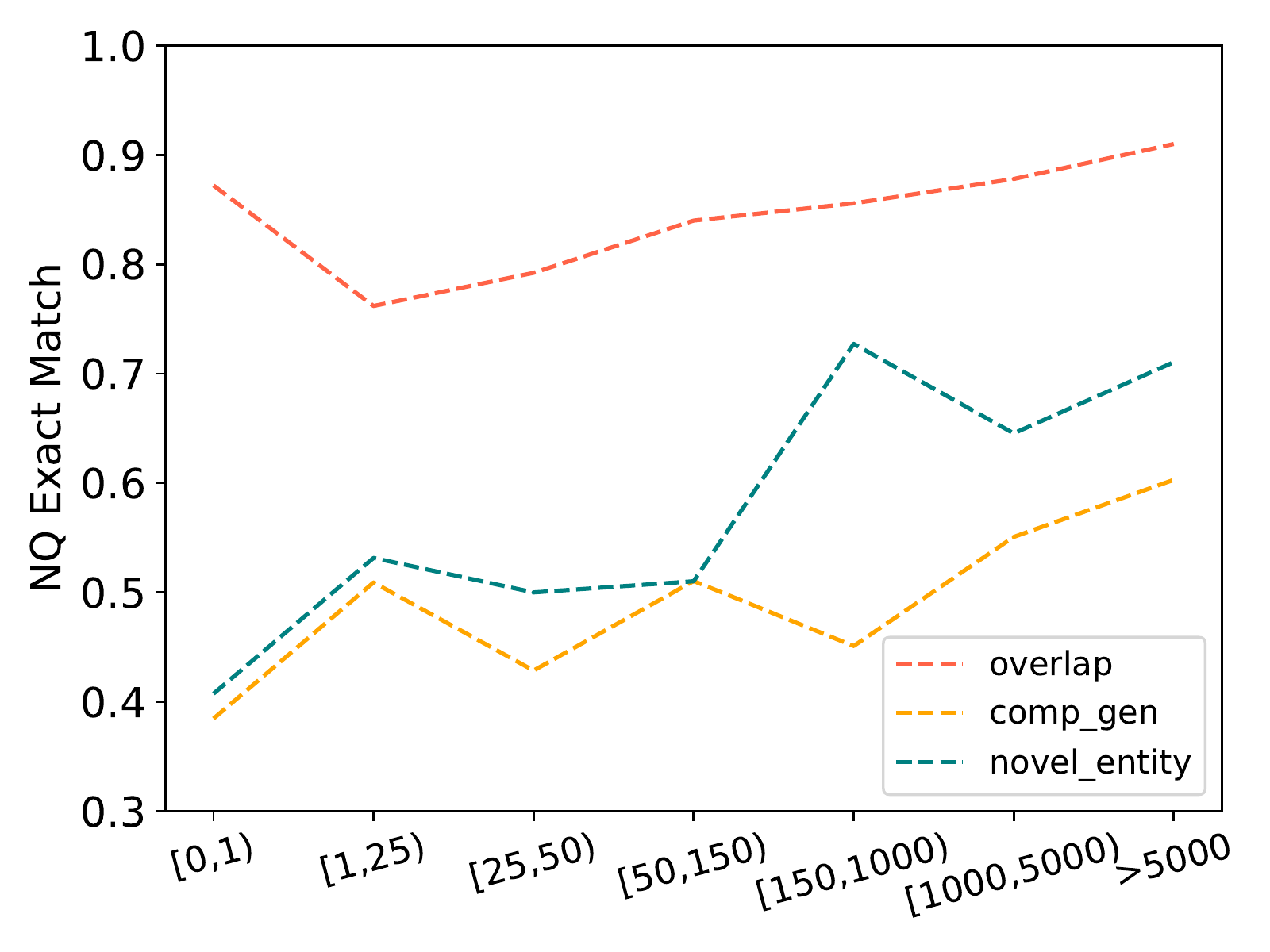}
}
\subfigure{
\includegraphics[width=.78\columnwidth]{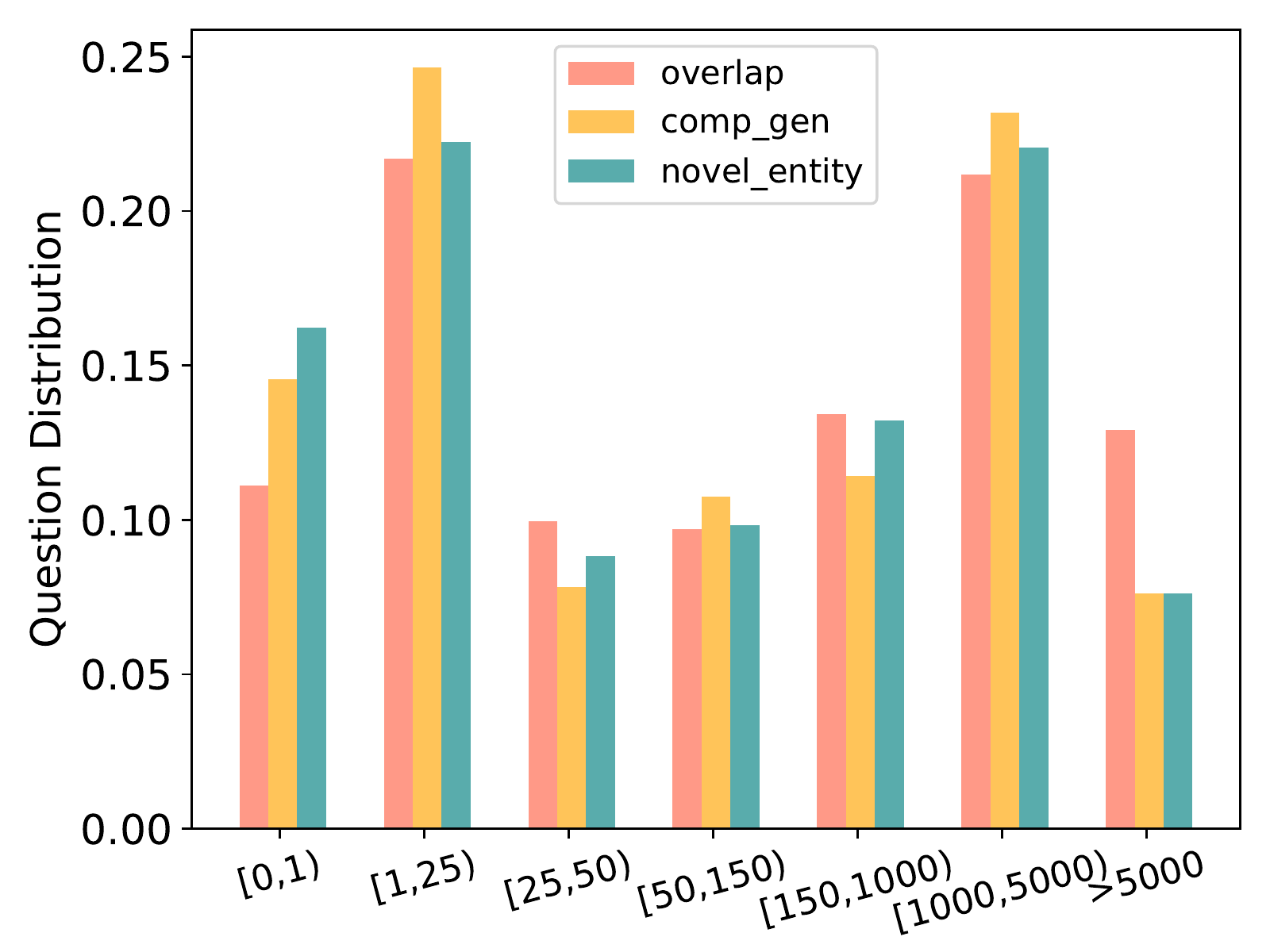}
}
\caption{
    Influence of question pattern frequency, where test questions are binned based on the frequency of their question pattern in the \textit{training set}.
}
\label{fig:question_pattern_freq}
\end{figure}

One might ask questions such as \textit{``Who plays the doctor in Sons of Anarchy?''} and \textit{``Who plays Stacey's mum in Gavin and Stacey?''}. Although semantically different, they share the structure ``who plays [entity] in [entity]'', which we refer to as a question pattern. To study if the frequency of these patterns affect model performance, we collect question patterns by replacing all wiki\_entities in a question with the token \textit{$\text{[entity]}$}, unifying the prepositions, and stemming each word. 

We group test questions for each category by the frequency of their patterns in the training set. 
In Figure~\ref{fig:question_pattern_freq}, we analyze FiD as an example since it achieves the highest EM score on unseen questions (results for other models can be found in Figure~\ref{fig:question_pattern_other_model} in the Appendix).
In the upper figure, the EM scores show that the model is more likely to make correct predictions for more common patterns.
Given this observation, we would like to investigate if the significant performance edge of the overlap category is due to a larger percentage of more frequent patterns. 
According to the lower figure, which shows the proportion of questions for each frequency bin,
the frequency distribution for each category is largely similar. Therefore the performance gap between overlap and the other two categories can not simply be explained by a difference in pattern distribution.

In Figure~\ref{fig:question_pattern_freq}, we also note that as the pattern frequency increases, the performance between comp-gen and \zeroshot{} diverges (for concrete question pattern examples see Figure~\ref{fig:pattern_example}).
This gap has a significant effect on overall model performance, since common patterns make up a majority of the test set.
Based on error analysis~(see Appendix~\ref{appendix:question_pattern} for details), we hypothesise that in the retrieved passages for comp-gen questions, answers do not always co-locate with the question anchor words.
This indicates future research should encourage the retriever to fetch passages that cover all aspects of the question in order for it to be answerable.
Under the assumption that the model could answer all patterns of questions equally well, regardless of frequency, the overall performance would be improved by $\sim 11\%$.
%

\newcommand{\tabfigure}[2]{\raisebox{-.5\height}{\includegraphics[#1]{#2}}}

\begin{figure}
\centering
\includegraphics[scale=0.4]{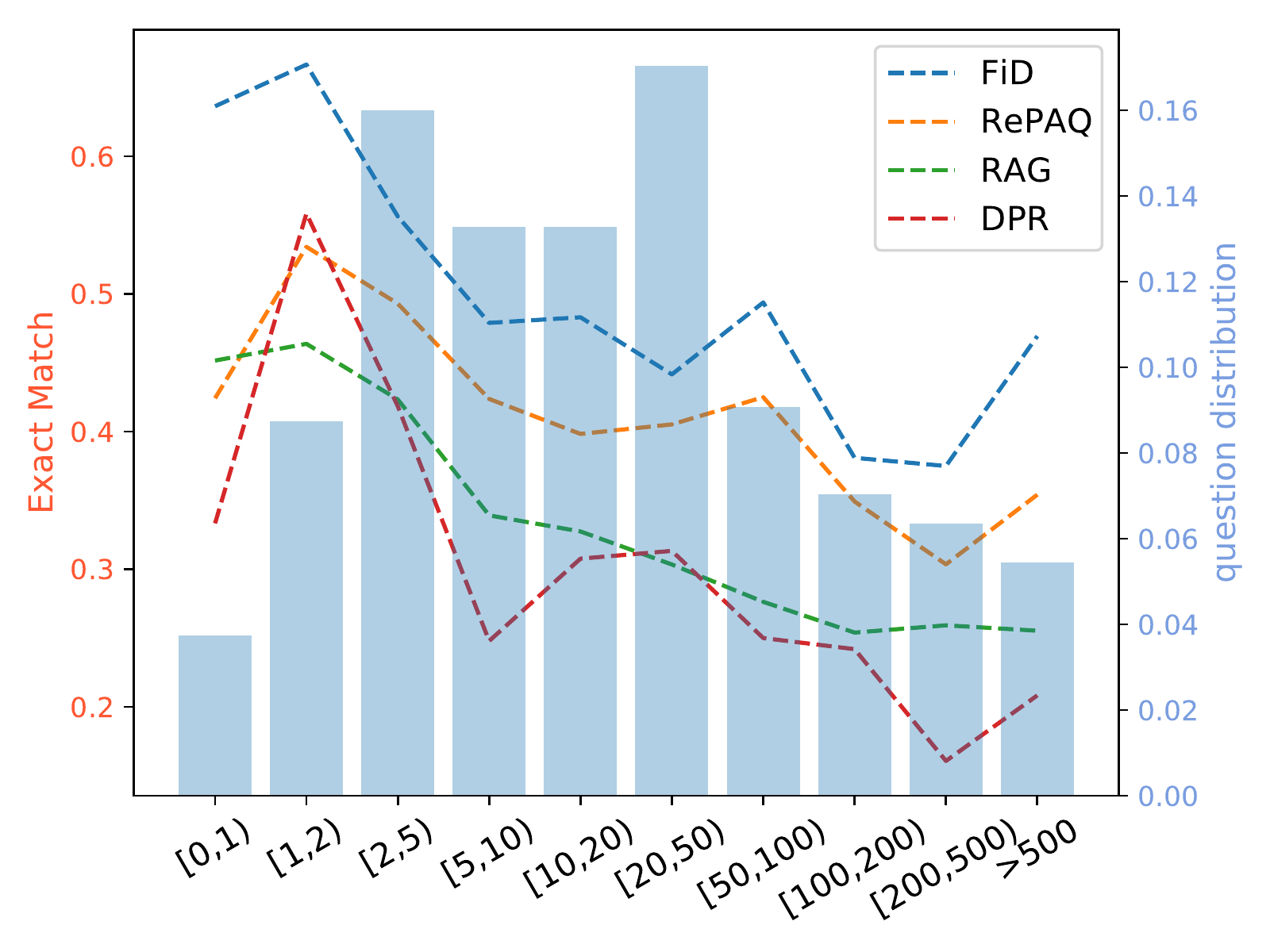}
\caption{
    Plot showing the influence of the wiki\_entities frequency in the question.
    The x-axis represents the wiki\_entities frequency in the training set and we use the most frequent wiki\_entities in each comp\_gen question.
}
\label{fig:wiki_ents_freq}
\end{figure}

\subsection{How do Non-parametric Models Handle Comp-gen Questions?}
\label{sec:entity_freq}
We use the decomposed atoms as the basis for our analyses on comp-gen questions.
Following the previous subsection \ref{sec:question_decomp}, we know that wiki\_entities leverage crucial semantics for factoid questions and Wikipedia is the most widely used source of knowledge in current ODQA datasets \cite{yang2015wikiqa,hewlett2016wikireading, rogers2021qa}.
Therefore, we would like to carefully study if the training set \textbf{wiki\_entities} frequency affects model performance.
Figure~\ref{fig:wiki_ents_freq} plots the EM score as a function of how often a test question's wiki entity appears in a training question.
We see that test accuracy is \emph{anti-correlated} with the training-set frequency of test questions' entities.
At first glance, this result seems surprising, and inconsistent with the well-known difficulty of modeling long-tail phenomena.
However, the following interpretation helps to explain this apparent contradiction.

We manually inspect the questions with the most frequent wiki\_entities, and find most of them are questions about countries, which is a frequent question topic in the NQ training set. %
For example, for the question \textit{``How many farmers are there in the USA''}, almost all the retrieved passages are highly relevant. 
The gold answer is ``3.2 million'' with the context \textit{``There were 3.2 million farmers''}. 
The model, however, generates the answer ``2.2 million'', taken from the context \textit{``There were 2.2 million farms\ldots''}.
Both passages come from an article titled ``Agriculture in the United States'', and the model is failing to draw a distinction between \textit{farms} and \textit{farmers}.
While it is easier to retrieve relevant documents for questions with more frequent wiki\_entities \cite{chen2021evaluating}, 
the passages retrieved for high-frequency entities are much more likely to contain type-consistent close-negatives and distractors, making it more difficult for the model to select the correct answer.
In other cases, questions are highly ambiguous, such as, ``\textit{What is the average salary for a US congressman}'', the gold answer \textit{\$174,000} applies for the year 2012, while predicted answer \textit{\$169,300} applies for the year 2008.
For NQ, the existence of high-frequency entities could be indicative of an ambiguous question. If we conduct an analysis using the NQ dev set annotations provided by \citet{min2020ambigqa}, we note that 50\% of questions with the entity ``\textit{US}'' and 64\% of questions with the entity ``\textit{NBA}'' are ambiguous.
To quantify the impact, using FiD as an example, we note that if we match the performance of comp-gen questions with common wiki\_entities to those with the unpopular wiki\_entities, the accuracy could be improved with $\sim 4\%$ points.

Besides wiki\_entities, it's prudent to consider the remaining atoms as well. The results are illustrated in Figure~\ref{fig:all_other_atom} and some findings are observed in the following:
1) For \textbf{question word}, all models achieve better performance for questions asking about WHO and WHICH, while performing worse on questions without any question word. Although EM scores drop significantly for WHY questions, it is hard to draw conclusions as there are only limited number of them in the test set. 
2) There is no clear correlation between model performance and \textbf{verb} frequency. Some of the ``best performing'' verbs are: sing, sang, wrote, and play, which closely correlate with the most frequent question patterns such as ``who sing song [ent]''. 
3) Since there is no clear correlation between model performance and other\_args frequency either, we group test questions based on the number of \textbf{other\_args} in each of the questions. 
It shows that models achieve higher EM scores on questions with fewer other\_args.
Interestingly, the most performing other\_args are closely related to WHO and WHICH questions, such as ``'s wife'', ``main character'', and ``tv show'', while the ``worse performed'' other\_args are mostly the comparative and superlative adjectives such as ``biggest house'' and ``second largest'' (also observed in \citealp{dua2019drop}). 

To summarize, the \textit{remaining atoms} are codependent on each other, especially for limited-length factoid questions. They should preferably be treated as a single unit (e.g. question pattern) to arrive the meaning of the question. In essence, their compositionality cannot be ensured and isolated \cite{dankers2021paradox}. 
Wiki\_entities on the other hands are independent of the context. The question is meaning-preserving even under wiki\_entities substitution.
The subpart for ODQA compositionality should focus on wiki\_entities and question patterns. As discussed above, their individual frequency have different impacts on the various components of ODQA models.

\subsection{How do Non-parametric Models Handle \capzeroshot{} Questions?}
\label{sec:zero_shot_ents}
Although we explicitly categorize unseen questions into comp-gen and \zeroshot{}, broadly speaking, questions with novel entities also require the model to generalize to novel compositions and thus could be considered to belong to the comp-gen category.
We seek to understand if the novel entities are the main bottleneck for ODQA models, or the model can handle them well enough to process the questions appropriately.
To explore this issue further, we run an ablation study, where, at inference time, we replace the novel entities in the question \emph{and} the support passages with an entity that has been seen from the training set.
Our experimental setup is working under the following constraints: 1) There can be only one wiki\_entity mentioned in the test question, so that replacing it will not risk altering the semantics of the original question. 2) The replacement entity must not be present in the original test question or its retrieved passages.

We run the inference for FiD model on 100 eligible questions, and find the model rarely changes its predicted answers, despite the modification, with 73\% of the predicted answers remaining unchanged.
We manually verified the remaining questions and observe that some differences are due to inherent limitations of our entity-swapping process, such as errors in entity-linking~(see Appendix~\ref{appendix:entity_swap} for examples).
Interestingly, we find that three altered questions give the right answers, despite originally generating incorrect ones.
Given these observations, we suggest that the model learns relatively good contextual embeddings for the novel entities by exploiting the context provided by the passages.
Thus, specific unseen entities are not the main bottleneck for the model to locate the desired answers.
\section{Related Work}

\subsection{Open Domain Question Answering}
Early systems relied on surface text pattern matching methods to detect answers \cite{ravichandran2002learning, soubbotin2001patterns}.
For traditional ODQA systems, linguistic experts first identify a set of question types and expected answer types using rule-based mapping methods for each type of questions \cite{allam2012question}. 
The input question needs to be classified into a certain type or taxonomy in order to be answered \cite{li2002learning, suzuki2003question}.
This approach is sub-optimal for most realistic use-cases, as it is not possible to enumerate all possible question types.

With the introduction of deep neural networks, recent ODQA system mostly adopt a ``Retrieve-and-Read'' architecture, popularized by \citet{chen2017reading}, retrieving relevant documents for a given question and inferring an answer from these documents.
Recent retriever models learn to encode questions and documents into dense vectors to score their similarity \cite{lee2019latent,karpukhin2020dense,khattab2021relevance}. 
Reader models can be categorized into \textit{extractive} models that predict an answer span within the document \cite{das2018multi, lin2018denoising, yang2019end} and \textit{generative} that generate answers condition on the question and the retrieved passages \cite{lewis2020retrieval,izacard2021leveraging}. 
Recent ODQA models provide substantial improvements over traditional systems \cite{zhu2021retrieving}, but as shown in Section~\ref{sec:question_pattern}, they still struggle with complex and infrequent questions.

\subsection{ODQA Model Analysis}
Retrieving relevant passages is an essential component for open-book ODQA models. 
A broad spectrum of recent work apply transformer~\cite{vaswani2017attention} models such as BERT~\cite{devlin2019bert} for information retrieval~\cite{yates-etal-2021-pretrained}. 
Following the success of using pretrained language models \cite{craswell2020overview}, studies have been made regarding 
their properties.
\citet{10.1162/tacl_a_00369} compare the lexical-matching abilities of these models to traditional methods such as BM25.
\citet{ma2021replication} and \citet{wang2021bert} study reproducibility, and demonstrate improvements by combining lexical-matching and dense retrievers.
\citet{thakur2021beir} introduce the BEIR benchmark to study zero-shot generalization for multiple neural retrieval approaches. 
Their conclusion is consistent with our findings that there is considerable room for improving the generalization of dense-retrieval models. 

To infer answers from retrieved documents, models generally use a \textit{reader} component implemented as a neural Machine Reading Comprehension~(MRC) model. 
Previous work has analyzed the MRC model by crafting adversarial attacks \cite{jia2017adversarial, mudrakarta2018did}, studying the difficulty of popular benchmarks \cite{kaushik2018much}, and demonstrating annotation bias \cite{gururangan2018annotation,sugawara2018makes,chen2019understanding}.
Despite the success for various datasets, there is little work analyzing the whole pipeline of question answering systems.
%
%
%
\citet{lewis2021question} showed that models perform substantially worse
on questions that cannot be memorized from training sets.
\citet{krishna2021hurdles} found that long-form question answering~(LFQA) systems do not ground their answers in the retrieved passages.
In contrast, for ODQA, we observe that when we replace retrieved passages with randomly-sampled passages at inference time, the model FiD \cite{izacard2021leveraging} largely fails to correctly answer any questions (see Appendix~\ref{appendix:attrfid} for experimental details).
%
%
\citet{gu2021beyond} define similar generalization levels based on schemas for Knowledge Base Question Answering.
However, our setting works without a schema and our generalization categories are derived from question decomposition atoms.
\section{Conclusion}
We study ODQA model generalization and categorize unseen questions into three subsets: \textit{overlap}, \textit{comp-gen}, \textit{\zeroshot{}}.
Treating questions as being compositional, we decompose them into atomic elements based on their semantics.
We believe that this decomposition strategy can help future work related to question structure and unification.
We evaluated several recent ODQA models on these three subsets for three popular datasets.
Our experimental findings both pinpoint the specific problems when handling different categories of novel questions and shed light on how to compositionally approach the factoid questions in ODQA task.
%


\bibliographystyle{acl_natbib}
\bibliography{acl2021}

\clearpage
\newpage
\appendix
\section{Appendix}
\subsection{Question Decomposition}
\label{appendix:question_decomp}

\newcommand{\qw}[1]{$\underline{\text{#1}}_{\textit{qw}}$}
\newcommand{\predicate}[1]{$\underline{\text{#1}}_{\textit{verb}}$}
\newcommand{\oarg}[1]{[\textit{other\_args}: #1]}
\newcommand{\wikient}[1]{[\textit{wiki\_entities}: #1]}

Below is a random selection of question decomposition examples from the NQ dataset.
In each question, \qw{x} denotes the question\_word, \predicate{y} denotes the verb, and the spans of other\_args and wiki\_ents spans are denoted by brackets.
%
%
Note that these structure slots are not always fully present in the question~(e.g, Q3, Q4, Q6, Q7, Q10).

As we rely on automated systems as a part of our decomposition process, this leads to the following limitations.
At times, the ELQ model fails to label wiki\_ents, such as for Q8 where \textit{every light in the house} is marked as other\_args.
Furthermore, as seen in Q9 there is the possibility of multiple question words being present although our approach only extracts a single question\_word.
Limitations such as these is one motivation for why we elected to perform manual verification for each question~(Section~\ref{sec:manual_verify}).
%

%
%
%

\begin{enumerate}
    \item \qw{Who} \predicate{is} the \oarg{owner} of \wikient{Reading Football Club}?
    \item \qw{Who} \predicate{died} in the \oarg{plane crash} \wikient{Grey's Anatomy}?
    \item \oarg{Cast} of \wikient{Law \& Order Special Victim Unit}?
    \item \qw{When} did \wikient{United States} \predicate{enter} \wikient{World War I}?
    \item \qw{Where} are most \wikient{nutrients} \predicate{absorbed} in the \wikient{human digestive tract}?
    \item \qw{When} did the \oarg{government} \predicate{change} the \oarg{retirement age}?
    \item \qw{What} \predicate{is} the \oarg{name} of the \oarg{gap} between \oarg{two front teeth}?
    \item \qw{Who} \predicate{sings} \oarg{every light in the house is on}?
    \item \qw{Where} \predicate{are} the \wikient{Winter Olympics} and when do they start?
    \item \wikient{Swan Lake} \wikient{the Sleeping Beauty} and \wikient{the Nutcracker} \predicate{are} \oarg{three famous ballets} by?
\end{enumerate}

\subsection{Question Collection for Human Verification}
\label{appendix:question_collection}
%
%
We use the following selection criteria to collect candidate questions for human verification.
For the overlap subset, as a first step, each $q$ is paired with each train question that shares the same answer or have answers which are a sub-sequence of $q$'s answer.
As a second step, we then require that the train question's similarity measurement score to $q$ is over a pre-defined threshold and that they have the same wiki\_entities as $q$.
%
%
%
For the remaining test questions, we consider $q$ as a candidate for comp-gen if all of its parsed elements are covered by the collection of all parsed elements in the training set. 
Lastly, if there exists any novel wiki\_entities in $q$ which are not present in the training set, $q$ is considered as a \zeroshot{} candidate.

\subsection{Generalization Subset Details}
\label{appendix:subset_detail}
As guidelines for the human annotators, we provide the following to resolve ambiguous or potentially problematic cases:
1) For overlap, we only consider questions that are superficial paraphrases and exclude those that require more complex forms of reasoning (e.g. \textit{Who played Mark on the show The Rifleman?} / \textit{Who played the boy on the show The Rifleman?}). 
2) For comp-gen, all other\_args in the test question must be covered in the collection of training set entities and all question\_word atoms alongside with the verb must be present in the training set.
However, there are questions where other\_args are not covered in the training set~(e.g. \textit{Animation Resort}) or are highly specific due to the decomposition processing and thus not covered~(e.g. \textit{fourth movie} compared to \textit{movie} or \textit{three different types} compared to \textit{types}) and are thus excluded from comp-gen.
3) For \zeroshot{},
%
%
there are cases when ELQ fails to extract wiki\_ents in questions because of words variation, such as \textit{Who sang It Going to Take a Miracle?} compared to the correct wiki\_ents \textit{It's Gonna Take a Miracle}.
4) There are also intrinsic problems in the datasets, some test questions are exactly the same as train questions but paired with different answers: (\textit{Where did Dolly Parton grow up?} with the answer \textit{Tennessee} and \textit{Where did Dolly Parton grew up} with the answer \textit{Sevierville}).
Following this manual verification, for Natural Questions, WebQuestions, and TriviaQA, 70.3\%, 81.3\%, and 69.5\% of their test questions are covered in the generalization subsets respectively.
%

%
%

%

\subsection{FiD Performance Analysis}
\label{appendix:implication_for_modeling}
Among the non-parametric models, FiD achieves the highest EM scores for both comp-gen and \zeroshot{} questions. 
We are interested in understanding if FiD's improved performance is due to leveraging a greater amount of contextual evidence provided by multiple passages, or whether it simply generates the most frequently-mentioned plausible answer.
We perform a simple experiment, by first collecting 544 questions answered incorrectly by FiD, where the gold answers occur less frequently than FiD's predicted answer in the retrieved passages. 
We then adjust the retrieved passages so that the original predicted answer and gold answer are mentioned an equal number of times, by masking out some of the original prediction mentions.
After adjusting the frequencies, we regenerate the answer predictions, and observe that FiD only produces 44 correct answers out of 544. 
This suggests that answer mention frequency is not the governing feature for FiD when generating answers on NQ.
It suggests the NQ FiD model adopts a strategy similar to a reranker, and extracts an answer from the highest latently-relevant document.

\subsection{Additional Question Pattern Analyses}
\label{appendix:question_pattern}
\begin{figure}
\centering
\includegraphics[width=0.9\columnwidth]{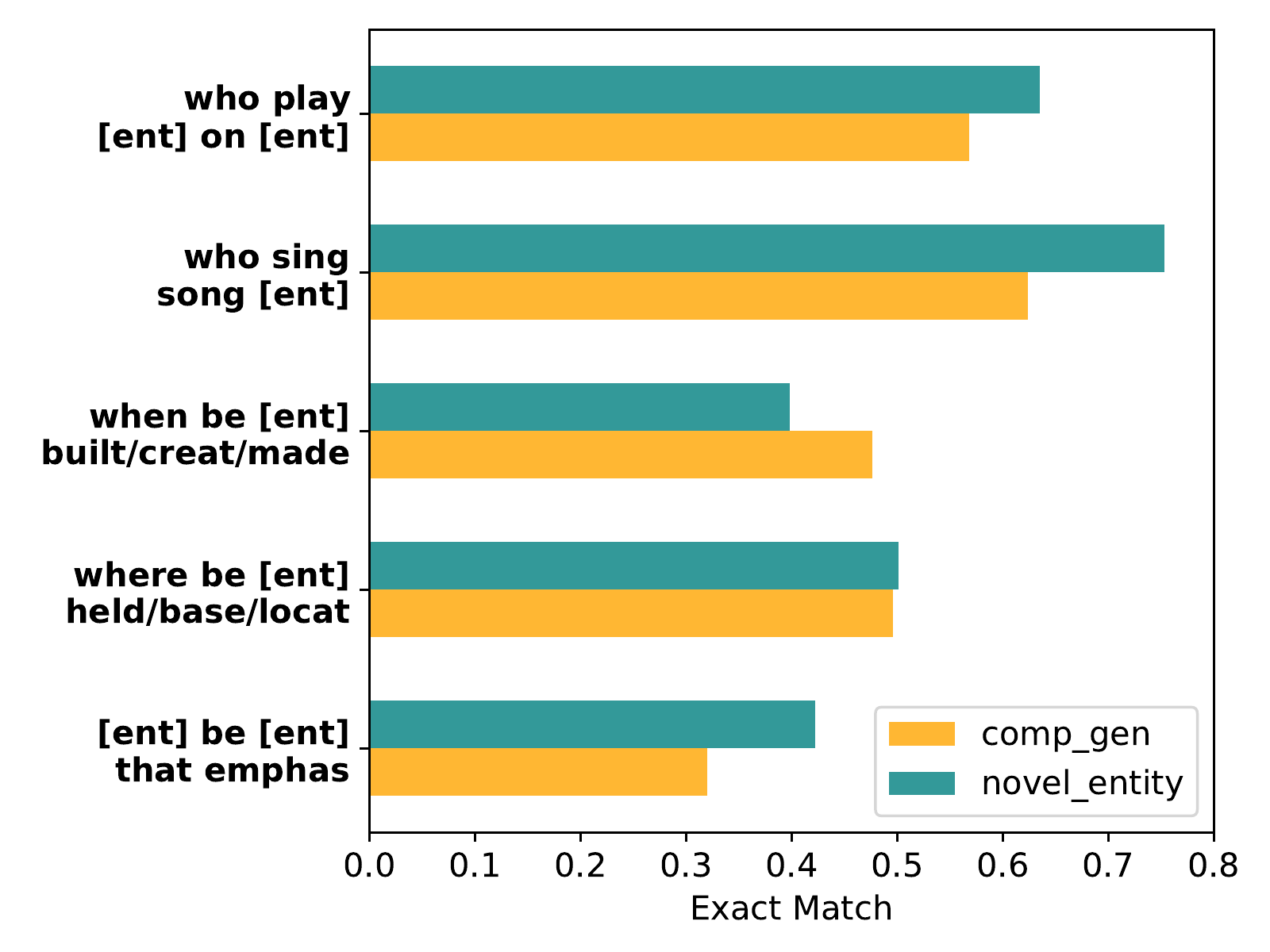}
\caption{
    Examples of question patterns and EM scores for their corresponding questions.
    For each question pattern, we sample the same number of comp-gen and \zeroshot{} questions.
    The two uppermost patterns are the most frequent~(thousands of occurrences), the following two are of medium frequency~(hundreds of occurrences), and the last is a novel pattern.
}
\label{fig:pattern_example}
\end{figure}

We sample the same number of comp-gen and \zeroshot{} questions for each example pattern, and display the results in Figure~\ref{fig:pattern_example}.
We checked several instances for the pattern ``who play [ent] on [ent]'', and find that the model fails more on comp-gen questions partially because the retrieved passages do not provide enough information to locate the answer. 
For example, for the question \textit{``Who played Mary in Christmas with the Kranks?''} none of the retrieved passages contain both \textit{Mary} and the movie name.
The model produces the answer \textit{Julie Gonzalo} from the passage \textit{Julie Gonzalo Julieta [...] is an [...] actress. [She] is also known for her roles ``Christmas with the Kranks''}, whereas the gold answer is \textit{Felicity Huffman} from the passage \textit{She also starred in [...] ``Christmas with the Kranks''}.
Since ``Mary'' is not mentioned in either passage, it is impossible to infer that the correct answer is \textit{Felicity Huffman}.
The support passages for \zeroshot{} questions, on the contrary, more often cover both of the anchor entities~(e.g. context \textit{Little Boy Blue is an ITV drama series ... Stephen Graham was cast as Detective ...} for the question \textit{``Who played the detective in Little Boy Blue}'').


\begin{figure*}%
    \centering
    {\includegraphics[scale=0.32]{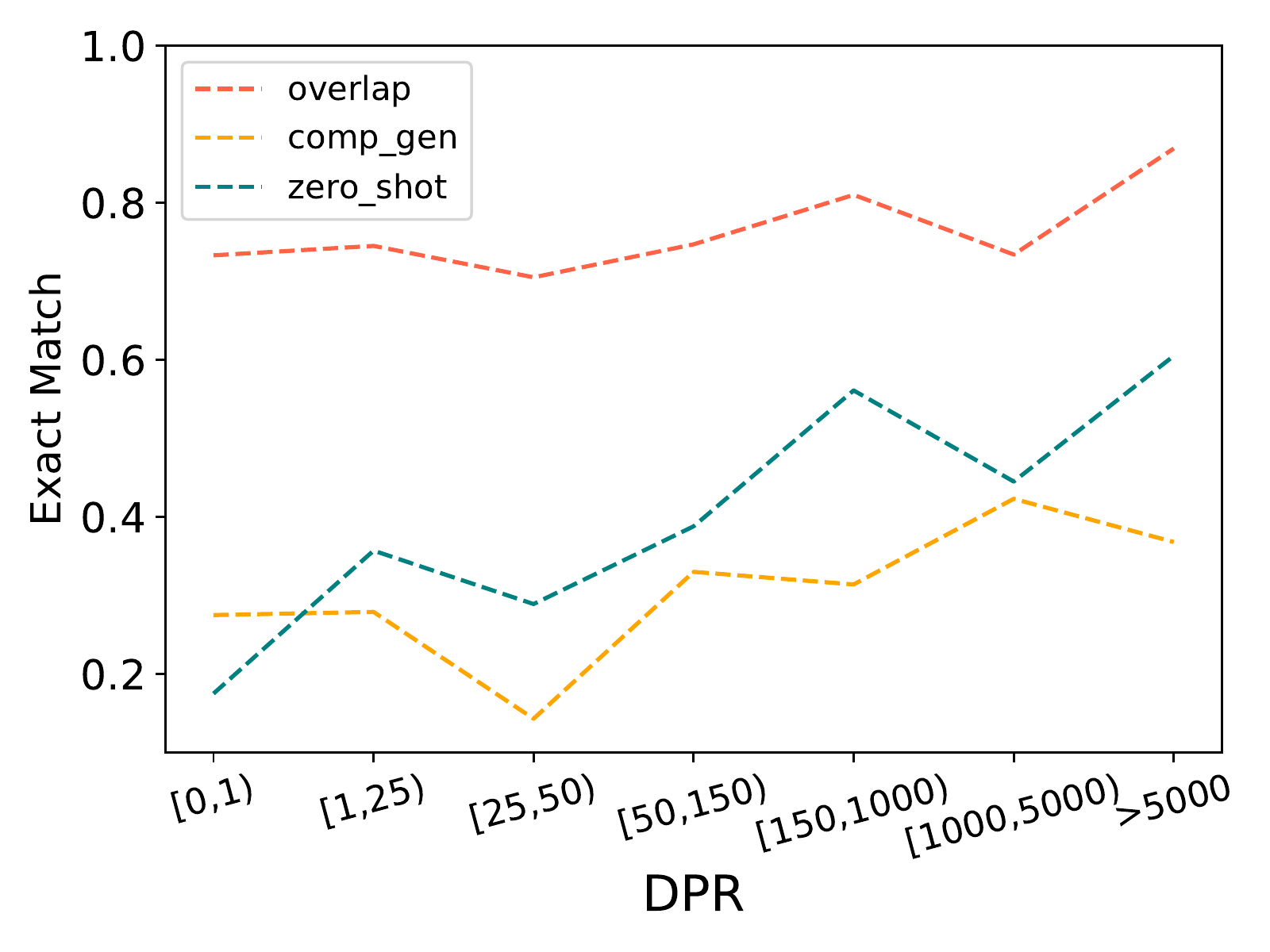}}%
    {\includegraphics[scale=0.32]{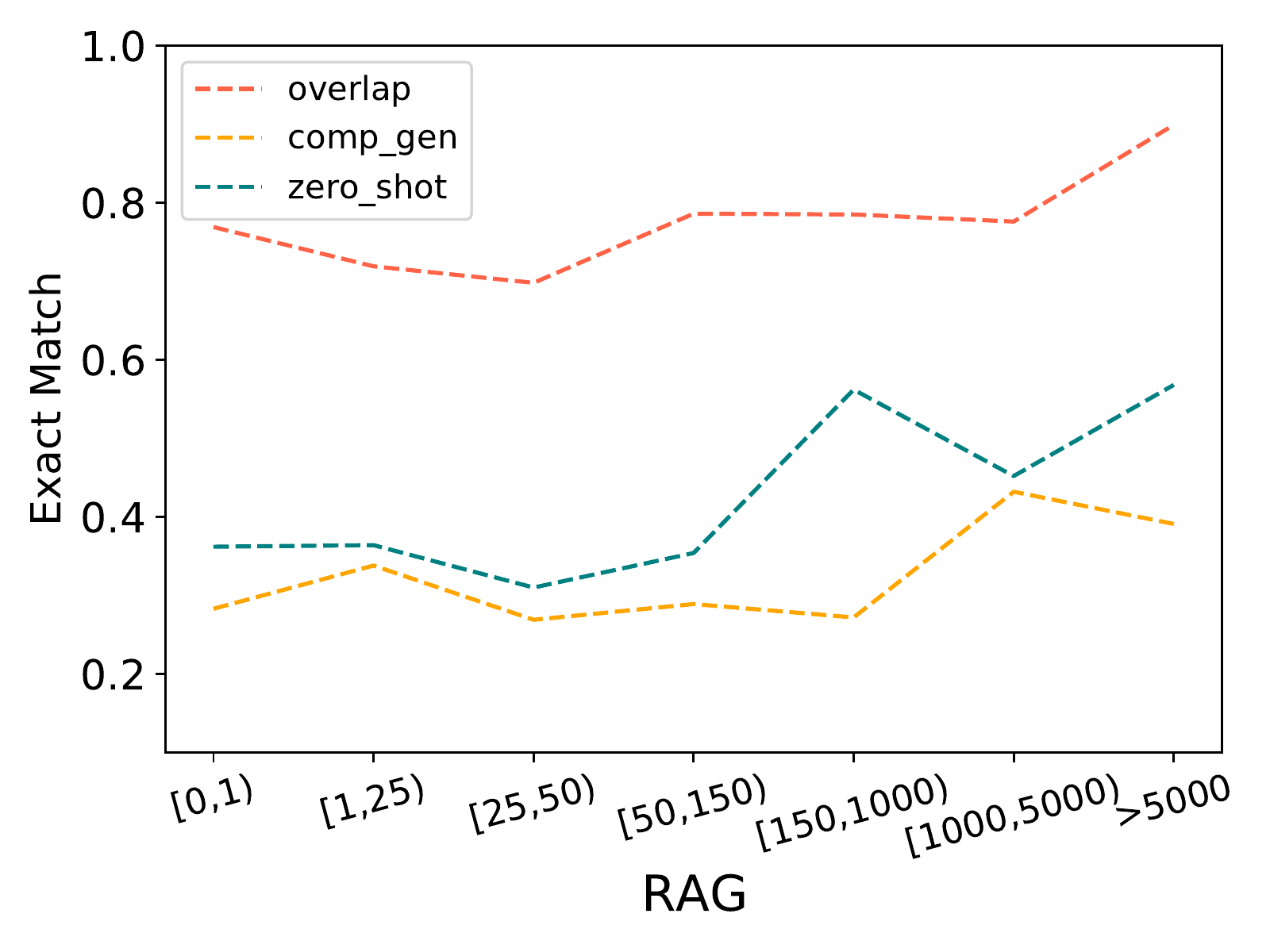}}%
    {\includegraphics[scale=0.32]{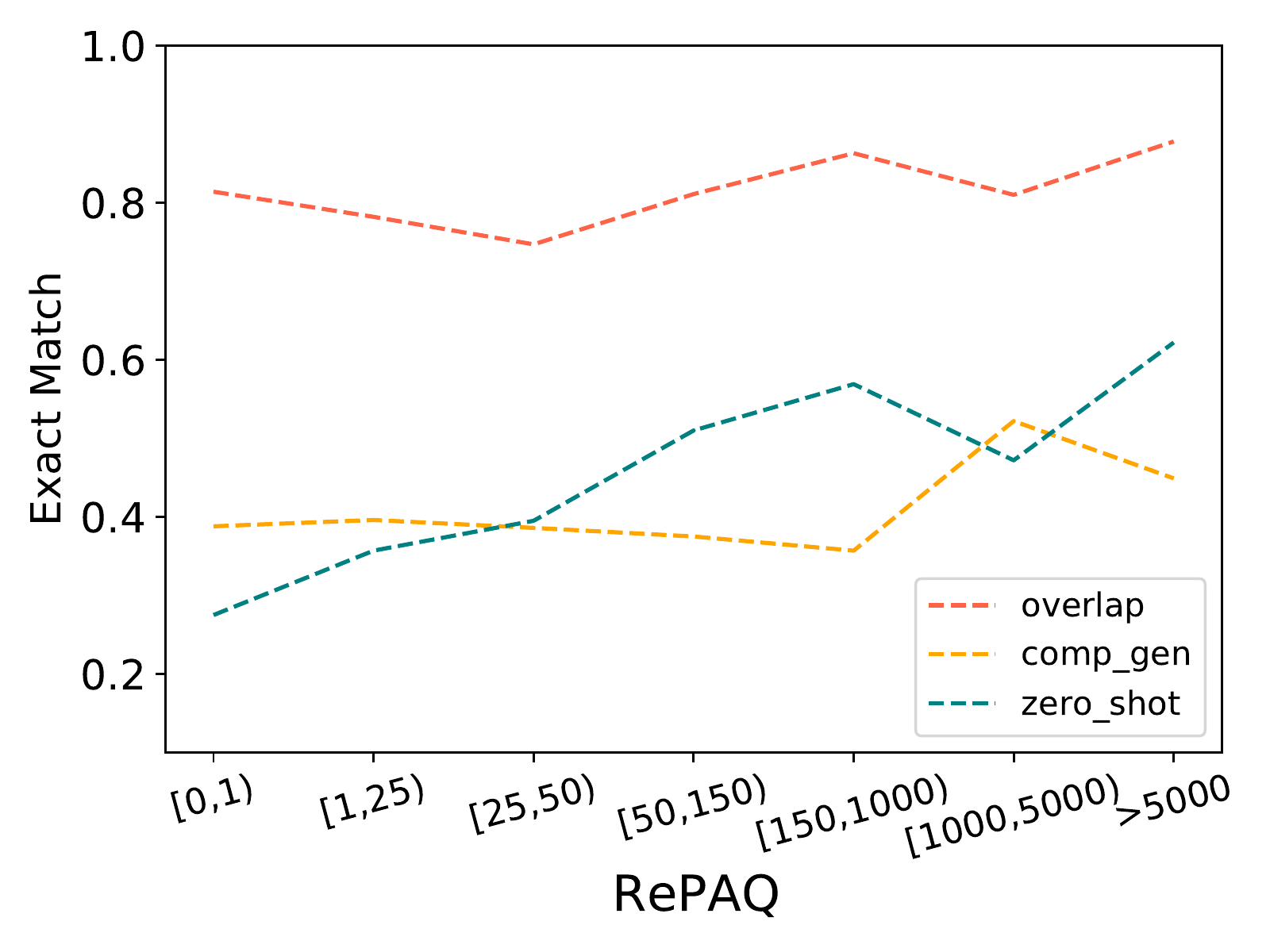}}%
    \caption{Influence of question pattern frequency. Each figure is associated with one non-parametric model, which is DPR, RAG and RePAQ from left to right. The test questions are binned based on the frequency of their question pattern in the training set.
    The y-axis shows the Exact Match score on the NQ test set.
    }%
    \label{fig:question_pattern_other_model}%
\end{figure*}

\begin{figure*}%
    \centering
    {\includegraphics[scale=0.32]{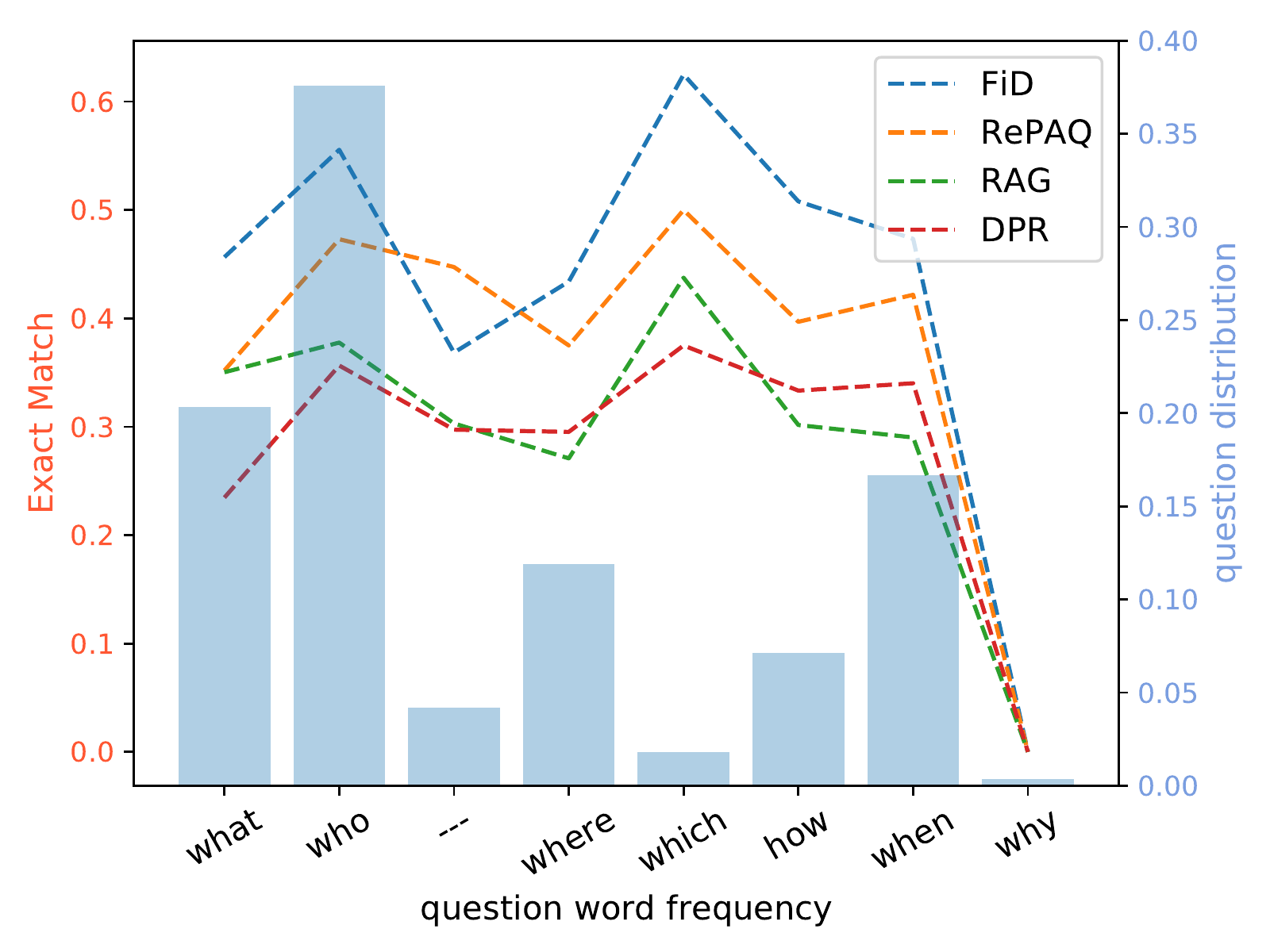}}%
    {\includegraphics[scale=0.33]{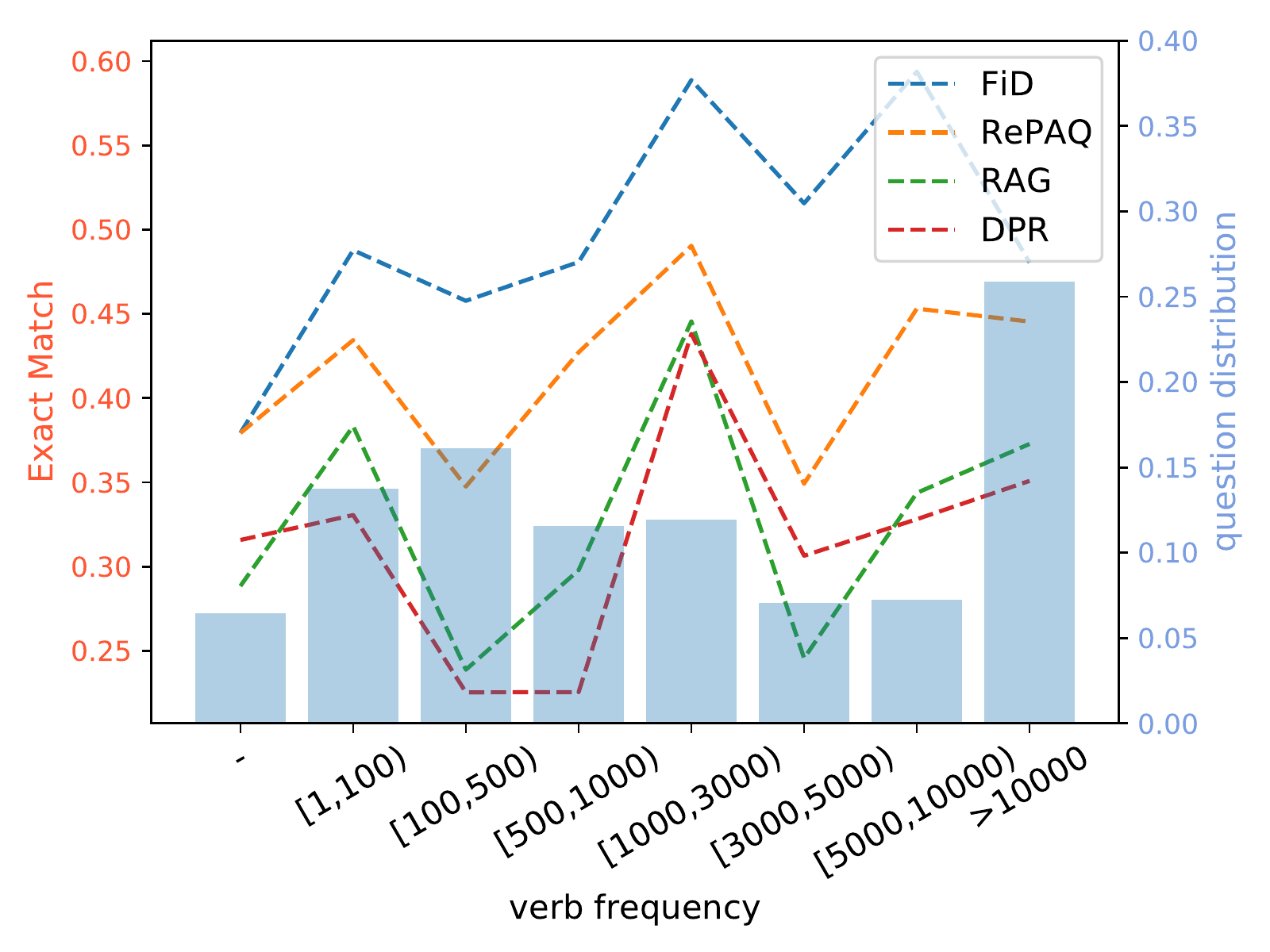}}%
    {\includegraphics[scale=0.31]{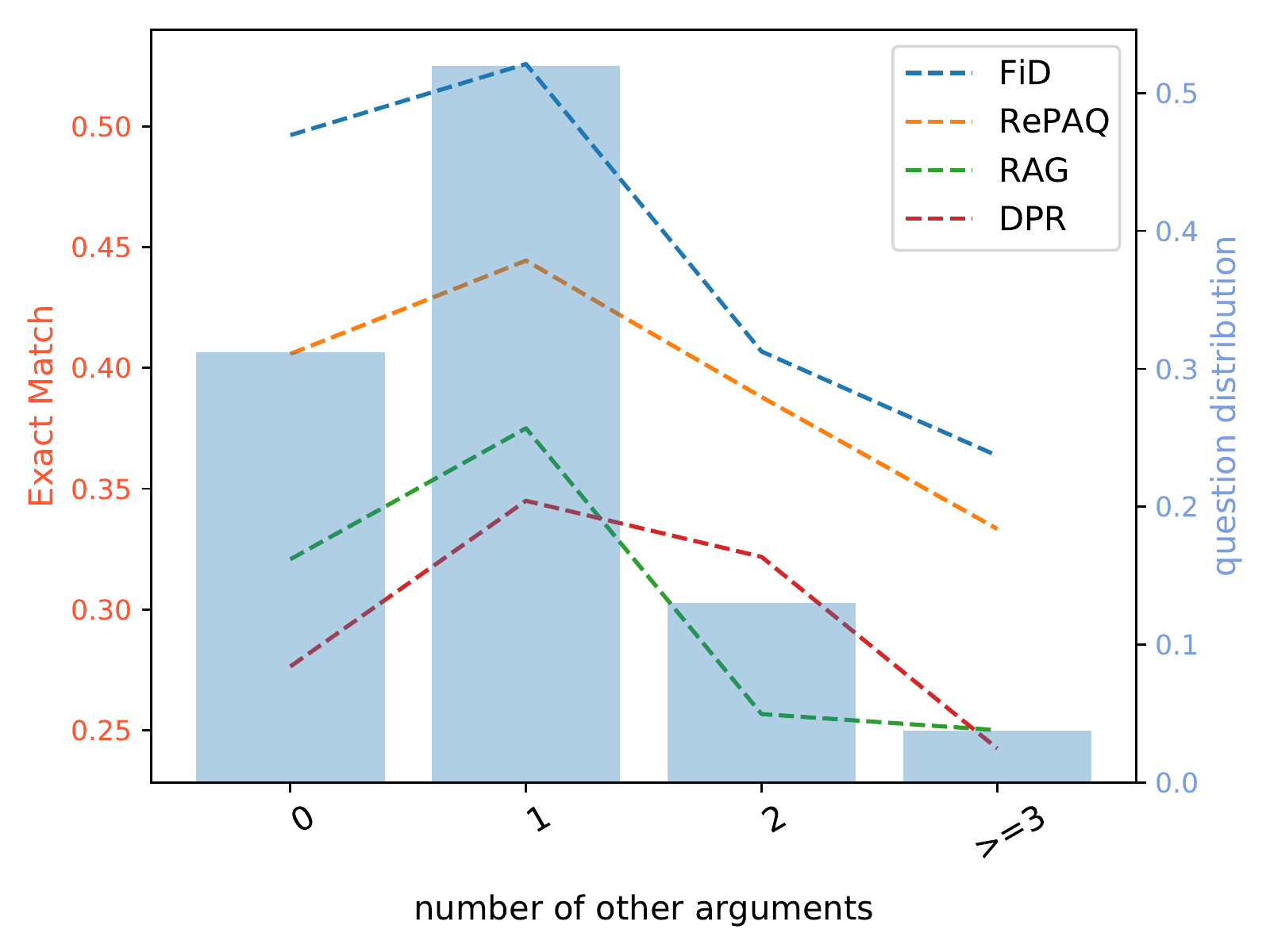}}%
    \caption{Influence of \textit{question word}, \textit{verb}, and \textit{other\_args} in the question~(from left to right).
    For the two left-most figures, the test questions are binned based on the individual atom frequency in the training set,
    ``-'' indicates test questions whose question word or verb is not covered in the training set.
    For the right-most figure, the x-axis shows the number of other\_args in each test question.
    All models are evaluated on the NQ test set.
    }%
    \label{fig:all_other_atom}
\end{figure*}

\subsection{Additional Non-parametric Generalization Analysis}
\label{appendix:entity_swap}


When analyzing the performance impact of the frequency of wiki\_entities in questions, one will have to account for the fact that there might be more than one entity present in the same question.
In our analysis in Section~\ref{sec:entity_freq} we consciously only considered the most frequent entity in a question. Note that we also experiment with the least frequent entity and they show the same negative correlation between entity frequency and performance.

%

%
As we noted in Section~\ref{sec:zero_shot_ents}, at times the novel entities in the original question may not match the corresponding mentions in the passage due to errors from the entity linking step.
For instance, for the question \textit{Who sings So Come and Dance with Me Jai Ho?} we swap the entity span ``So Come and Dance with Me Jai Ho'', however, this span is too wide as an entity as the correct entity would be ``Jai Ho''.
Therefore the model is unable to match the correct song name in the passage; thus giving a different answer.
Other error cases can be attributed to the granularity of the predicted answer: e.g. ``624 CE'' and ``13 March 624 CE''.
We do however note that for the great majority of cases our entity-swapping procedure works as intended.

\begin{table}
\centering
\resizebox{\columnwidth}{!}{
\begin{tabular}{lcccc}
\toprule
\textbf{Passage Processing} & \textbf{Total} & \textbf{Overlap} & \textbf{Comp-gen} & \textbf{\capzeroshot{}}  \\
\midrule
Original retrieved              & 53.1           & 78.9             & 40.0              & 47.7                     \\
50\% random                 & 53.2           & 78.3             & 39.9              & 48.3                \\
99\% random                 & 55.5           & 74.3             & 46.1              & 54.0                \\
100\% random                & 3.6            & 5.1              & 2.0               & 3.0                 \\
\bottomrule
\end{tabular}}
\caption{
    Comparison of FiD's predictions for the NQ test set, conditioned on the \textit{originally retrieved} passages and a gradually increasing number of \textit{randomly chosen} passages.
    x\% means the percentage of retrieved passages are replaced with random ones.
    For \textit{99\% random}, the rest passage is gold passage which contains the gold answer span.
}
\label{tab:random_psg}
\end{table}

\subsection{Answer Grounding in Retrieved Passages}
\label{appendix:attrfid}
We noted in Section~\ref{sec:discuss} that we find evidence the FiD~\citep{izacard2021leveraging} ODQA model does ground its answers in the retrieved passages.
This observation can be contrasted to that of \citet{krishna2021hurdles}, who found that answers to long-form questions were not grounded in the passage, in that models would provide the same answer regardless of the context provided.
A complete picture of the results from our experiment can be seen in Table~\ref{tab:random_psg}.
We note that when the models is fed solely random passages it fails to answer nearly all questions~(3.6\%).
However, but provided with half gold and half random passages, it performs on par with its original performance.
Lastly, we note that when presented with a single gold passage and otherwise only random passages, the model is still able to determine which passage is the gold passage and answer the question correctly -- in fact, the performance even improves upon the original performance with more than more than 5\% for comp-gen and \zeroshot{} questions.
%


%

%


%
        %

%

%

\subsection{Additional Examples for three generalization subsets}
\label{appendix:question_examples}
Additional examples from Natural Questions are provided in Table~\ref{tab:annot_NQ}, WebQuestions in Table~\ref{tab:annot_TQA}, and TriviaQA datasets in Table~\ref{tab:annot_WebQ}.

\begin{table*}
\resizebox{\textwidth}{!}{
\begin{tabular}{lll}
\toprule
\multicolumn{1}{l}{Group}  & Test question         & \begin{tabular}[c]{@{}l@{}} Train question\end{tabular}  \\ \hline
\multirow{4}{*}{Overlap}   &   Where does patience is a virtue come from           & Where did the saying patience is a virtue come from \\
& Who was the killer in the movie I Know What You did Last Summer & Who was the murderer in I Know What You did Last Summer \\ 
& When was the last time Arsenal win Premier League & When was the last time Arsenal won the Premier League title \\
& Where does blood go when it leaves the pulmonary artery & Where does blood go after the pulmonary artery \\
\hline

\multirow{4}{*}{Comp-gen}  &What is the most popular religion in Sweden        & What is the most popular religion in Ukraine\\
& What are the main functions of the stem  &  What are the main functions of the control bus  \\ 
& Who is in charge of ratifying treaties in the US & Who is in charge if president is impeached \\
& Cast of the Have and Have Nots play & The last episode of the Haves and Have Nots \\
\hline

\multirow{4}{*}{\capzeroshot{}} & Where does wild caught \textit{sockeye  salmon} come from   & When was \textit{Sony walkman} first sold in stores\\

& The probability of making a \textit{Type I Error} when retaining .. is & When was \textit{tower of terror} built in Disneyland \\

& Who was the \textit{Pinkerton Detective Agency} ’s first female detective & Who played \textit{detective Green} on Law \& Order \\

& Where was the \textit{world economic forum} held this year & Who holds the \textit{world record} for 100 meters \\

\bottomrule                      
\end{tabular}}
\caption{
    Example questions from NQ test set.
    }
\label{tab:annot_NQ}
\end{table*}

\begin{table*}
\resizebox{\textwidth}{!}{
\begin{tabular}{lll}
\toprule
\multicolumn{1}{l}{Group}  & Test question         & \begin{tabular}[c]{@{}l@{}} Train question\end{tabular}  \\ \hline
\multirow{4}{*}{Overlap}   &   Which is the highest waterfall in the world & What is the tallest waterfall in the world \\
& In the cartoon series, what kind of dog is Scooby Doo & What breed of dog is Scooby-Doo \\
& Who directed the film ``Gladiator'', starring Russell Crowe & Who directed the film Gladiator \\
& Which is the largest island in Canada & What is Canada's largest island \\
\hline

\multirow{2}{*}{Comp-gen}  & - What nationality was the painter Vincent Van Gogh        & - What nationality was painter Piet Mondrian\\

& \begin{tabular}[c]{@{}l@{}}- What post was held by Winston Churchill during \\ the 1926 general strike in the UK\end{tabular} & \begin{tabular}[c]{@{}l@{}}- What role was played by Arthur Cook \\ In the general strike of 1926\end{tabular} \\
&- By population, which is the second biggest city in France & \begin{tabular}[c]{@{}l@{}}- In terms of population, which is the \\ second largest city in Finland 1926\end{tabular}  \\
& \begin{tabular}[c]{@{}l@{}}- In humans, the medical condition prepatellar bursitis \\affects which part of the body\end{tabular} &
\begin{tabular}[c]{@{}l@{}}- The medical condition aerotitis affects \\which part of the human body\end{tabular}  \\
\hline

\multirow{4}{*}{\capzeroshot{}} & \begin{tabular}[c]{@{}l@{}} - In \textit{‘follow that camel’}, the fourteenth carry on film, \\ sid james was replaced by which us actor \end{tabular} & \begin{tabular}[c]{@{}l@{}}- What was the cause of death of carmen \\in the opera \textit{of that name} \end{tabular}
\\
& \begin{tabular}[c]{@{}l@{}} - Who has recently overtaken \textit{brian o'driscoll} \\ to become ireland's most capped player\end{tabular} & 
\begin{tabular}[c]{@{}l@{}} - In the 2005 remake of king kong, \\who played the writer \textit{jack driscoll} \end{tabular}\\
& - \textit{Shining Tor} is the highest point in which county & - \textit{Shinto} is the main religion in which country \\
& - Who had a \textit{Too Legit to Quit} tour & \begin{tabular}[c]{@{}l@{}} - Which sweets were advertised as \\ the \textit{Too Good to Hurry Mints} \end{tabular} \\

\bottomrule                      
\end{tabular}}
\caption{
    Example questions from TriviaQA test set.
    }
\label{tab:annot_TQA}
\end{table*}

\begin{table*}
\resizebox{\textwidth}{!}{
\begin{tabular}{lll}
\toprule
\multicolumn{1}{l}{Group}  & Test question         & \begin{tabular}[c]{@{}l@{}} Train question\end{tabular}  \\ \hline
\multirow{4}{*}{Overlap}   &   What is the currency of Puerto Rico called   & What type of currency is used in Puerto Rico \\
& Which countries speak German officially & What countries speak German as a first language \\
& What language is spoken in Haiti today & What language do Haitian speak \\
& What team is Hank Baskett on 2010 & What team is Hank Baskett playing for in 2010 \\

\hline

\multirow{4}{*}{Comp-gen}  & What year was George W Bush elected & What is George W Bush's middle name \\
& What year did the Seahawks win the Superbowl & In what Super Bowl did the Seahawks face the Steelers \\
& Where did Queensland get its name from & From where did the Guillotine get its name \\
& Where was Theodore Roosevelt buried & Where is George v1 buried \\
\hline

\multirow{4}{*}{\capzeroshot{}} & Where did \textit{Andy Murray} started playing tennis &  When did \textit{Sean Murray} first appear on NCIS\\
& What time in \textit{Hilo Hawaii} & Who was \textit{Phil Harris} married to \\
& Where did \textit{Bristol Palin} go to school & What team is \textit{Chris Paul} on \\
& What time does \textit{American Horror Story} air & Who made the \textit{American Red Cross} \\
\bottomrule                      
\end{tabular}}
\caption{
    Example questions from WebQ test set.
    }
\label{tab:annot_WebQ}
\end{table*}

\end{document}